\documentclass[11pt]{article}

\PassOptionsToPackage{hyphens}{url}
\usepackage[preprint]{acl}

\usepackage{times}
\usepackage{latexsym}

\usepackage[T1]{fontenc}

\usepackage[utf8]{inputenc}

\usepackage{microtype}

\usepackage{inconsolata}

\usepackage{graphicx}

\usepackage{algorithm}
\usepackage{algorithmic}

\usepackage{newfloat}
\usepackage{listings}
\DeclareCaptionStyle{ruled}{labelfont=normalfont,labelsep=colon,strut=off}
\floatstyle{ruled}
\newfloat{listing}{tb}{lst}{}
\floatname{listing}{Listing}

\usepackage{booktabs}

\usepackage{amsmath}
\usepackage{amssymb}
\usepackage{multirow}
\usepackage{tabularx}
\usepackage{enumitem}
\usepackage[most]{tcolorbox}
\usepackage{fvextra}
\usepackage{rotating}
\usepackage{siunitx}
\usepackage{array}
\newcommand{\hi}[1]{\textcolor{green!60!black}{#1}}
\newcommand{\lo}[1]{\textcolor{red!70!black}{#1}}

\newcolumntype{L}[1]{>{\raggedright\arraybackslash}p{#1}}
\newtcolorbox{examplebox}[1]{
    colback=gray!5!white,
    colframe=gray!75!black,
    fonttitle=\bfseries,
    title=#1,
    arc=1mm,
    boxrule=0.5pt,
    left=2mm,
    right=2mm,
    top=2mm,
    bottom=2mm,
    breakable
}

\newcommand{\rqOne}{\textbf{RQ1}}
\newcommand{\rqTwo}{\textbf{RQ2}}
\newcommand{\rqThree}{\textbf{RQ3}}
\newcommand{\personTerm}{\textit{Person}}
\newcommand{\situationTerm}{\textit{Situation}}
\newcommand{\behaviorTerm}{\textit{Behavior}}

\title{From Representations to Behaviors: Exploring the \personTerm--\situationTerm--\behaviorTerm{} Triad in LLMs}

\author{
  Ruikang Zhang\textsuperscript{\rm 1},
  Shuo Wang\textsuperscript{\rm 2}\thanks{Work done during internship at Peking University.},
  Qi Su\textsuperscript{\rm 1}\thanks{Corresponding author.}
\\
  \textsuperscript{\rm 1}Peking University, Beijing, China\\
  \textsuperscript{\rm 2}Beijing Institute of Technology, Beijing, China\\
  2300018416@stu.pku.edu.cn, 3120265734@bit.edu.cn, sukia@pku.edu.cn
}

\begin{document}

\maketitle

\begin{abstract}
Human personality theories characterize traits not as isolated attributes captured by a single score, but as stable individual tendencies expressed through the interplay among persons, situations, and behaviors. Existing studies of personality-related behavior in LLMs have primarily focused on outputs elicited under personality conditioning, characterizing observable trait-related expressions while lacking mechanistic evidence for the existence of internal personality-related representations, their cross-situational expression, and how these representations shape specific behaviors. Building on Funder's personality triad framework, we adapt its three components for LLM analysis: \personTerm~as personality-related internal representations, \situationTerm~as contexts that afford trait-relevant responses, and \behaviorTerm~as response patterns on broader social tasks. We introduce a framework for discovering, controlling, and validating trait-like representations in LLMs. First, using contrastive behavior pairs grounded in shared situations, we identify sparse internal features associated with opposing poles of personality traits through sparse autoencoder decomposition. We validate their trait relevance through effects on behavior to situation, token-level activation patterns, and robustness to paraphrasing. Second, feature-level interventions induce bidirectional trait-related shifts across a separate, diverse set of situations while preserving response validity, demonstrating consistent expression across contexts. Third, applying the same interventions to social intelligence tasks reveals behavioral changes with benefit-tradeoff patterns consistent with findings from human personality research, providing behavioral-level validation beyond personality scores. Our findings provide evidence that LLMs contain controllable trait-like representations linking internal states, situational expression, and behavioral outcomes. 
\end{abstract}

\begin{figure*}[t]
    \centering
    \includegraphics[width=\linewidth]{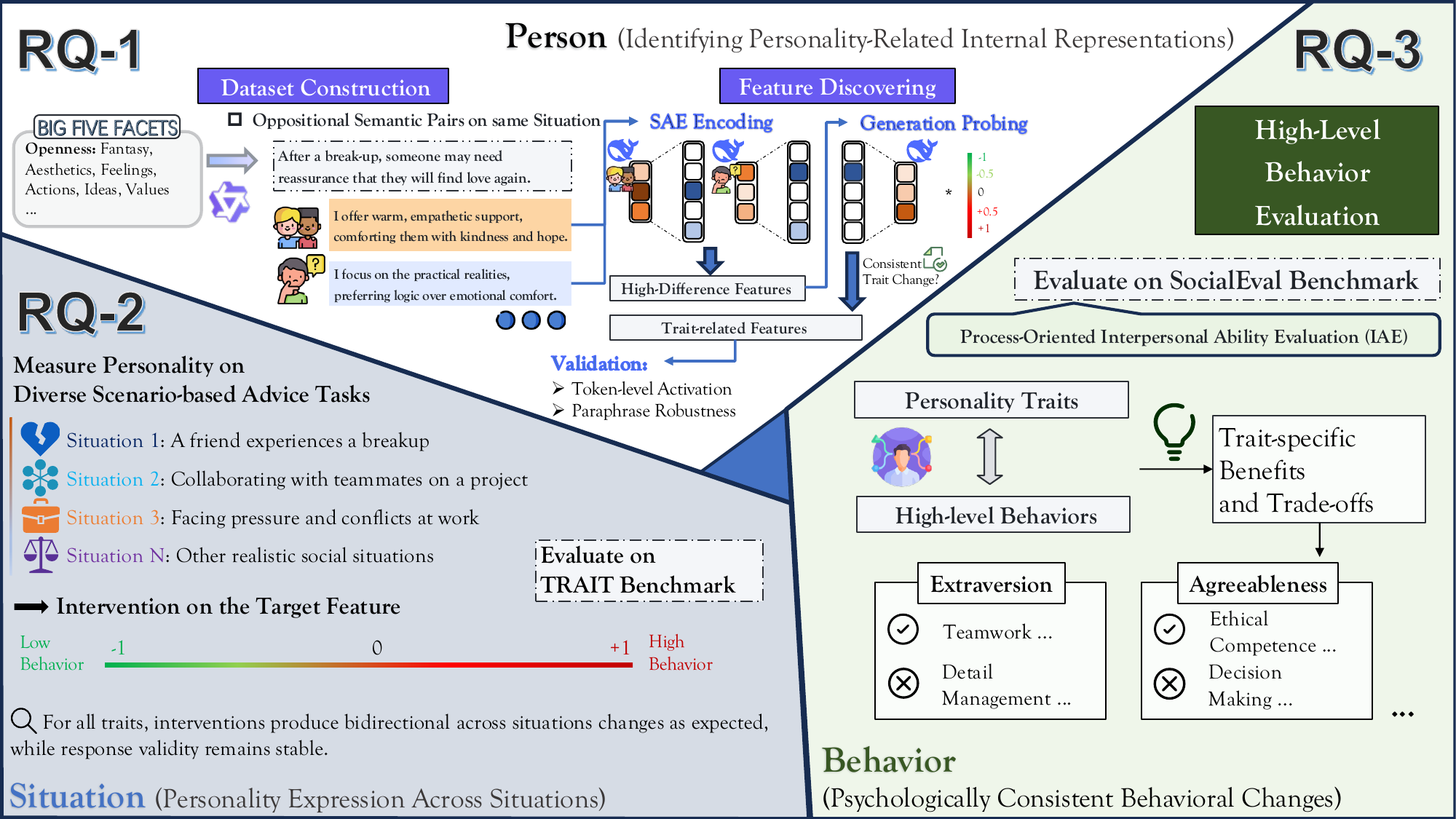}
    \caption{Overview of the study. Matched high--low behavioral contrasts under shared situations retrieve candidate SAE features, which are selected through generation-based intervention probing and characterized through token-level activation and paraphrase robustness (\rqOne--\personTerm). Interventions on the selected features evaluate bidirectional trait-relevant expression and response validity across a separate, diverse set of TRAIT situations (\rqTwo--\situationTerm). The same interventions reveal broader changes, including trait-specific benefits and trade-offs, across SocialEval interpersonal abilities (\rqThree-\behaviorTerm).}
    \label{fig:overview}
\end{figure*}

\section{Introduction} \label{sec:intro}

Personality is inferred from the expression of relatively stable individual tendencies across situations \cite{allport1937personality}. Human personality theory therefore treats persons, situations, and behaviors as jointly shaping action \cite{FUNDER200621}. This perspective is increasingly relevant to large language models (LLMs), where personality induction \cite{10.1007/978-981-97-9434-8_19} supports applications ranging from safety alignment, toxicity and bias analysis \cite{zhang2024betterangelsmachinepersonality,wang-etal-2025-exploring-impact}, and personalization to demographic and social simulation \cite{Cui2025}.

LLM personality can be induced through persona prompting \cite{chen2024from}, training and weight-space editing \cite{brito2025modeling,ye2026largelanguagemodelpsychometrics}, and interventions on internal activations \cite{chen2025personavectorsmonitoringcontrolling,ICLR2025_d399b67f,ICLR2025_25203d1c}. Evaluation has developed in parallel, progressing from direct Big Five and MBTI inventories \cite{Serapio-Garcia2025,cui2024machinemindsetmbtiexploration,song2023largelanguagemodelsdeveloped} to personality-profile emulation \cite{wang2025evaluating}, scenario-grounded choices \cite{lee-etal-2025-llms}, and open-ended linguistic assessment \cite{10.1162/coli_a_00550}. These approaches measure personality from closed-ended, single-option responses, option-token probabilities, or trait estimates derived from generated text. Mechanistic studies further identify personality-related directions, neurons, and attention heads, extending this broader study of model behavior of machine psychology \cite{hagendorff2024machinepsychology}.

However, these lines of evidence are rarely connected. On the representation side, a direction retrieved from contrastive prompts may primarily reflect lexicosyntactic regularities of the probing data, obscuring a more abstract personality-related signal. On the intervention side, persona prompts provide discrete external conditioning, whereas dense mean-difference vectors can strongly perturb generation; both complicate the relation between a manipulated internal state and coherent, instruction-following situated responses. On the evaluation side, inventory scores and option probabilities quantify whether responses align with the target trait, but coherent, instruction-following situational responses and broader social behavior require separate evidence, as persona conditioning can shift self-reported tendencies while leaving situational behavior weakly aligned \cite{han2025personality}. The unresolved question is therefore whether an internal representation associated with a personality trait can be traced from personality-related activation, through effective trait expression across situations, to systematic behavioral consequences.

We organize this problem through the \personTerm--\situationTerm--\behaviorTerm{} components of the personality triad framework. For LLMs, \personTerm{} denotes personality-related internal representations, \situationTerm{} denotes concrete contexts that elicit personality-dependent responses, and \behaviorTerm{} denotes observable behavioral manifestations in broader social tasks beyond direct personality assessment. This yields three research questions: (\rqOne--\personTerm) Can personality-related internal representations be identified from contrasting behavioral expressions under matched situations, beyond lexicosyntactic cues? (\rqTwo--\situationTerm) Does intervention on the identified representations elicit trait-relevant tendencies in both directions across a separate, diverse set of situations while preserving coherent, instruction-following responses? (\rqThree--\behaviorTerm) Does the intervention induce consistent changes across social behaviors that correspond to established findings in human personality research?

We study these questions using the Big Five \cite{goldberg1990alternative} and sparse autoencoders (SAEs) \cite{shu-etal-2025-survey}. First, we hold a situation fixed while constructing opposing trait-relevant reactions, and use their SAE activation differences for feature retrieval. Intervention probing then selects candidates with coherent, trait-aligned causal effects, while token-level activation analysis and paraphrase tests further validate the selected features. Second, we intervene on the selected features in a scenario-grounded personality benchmark, TRAIT \cite{lee-etal-2025-llms}, evaluating trait direction and response validity across a separate, diverse set of situations. Third, we apply the same interventions to a social-intelligence benchmark, SocialEval \cite{zhou-etal-2025-socialeval}, to examine broader behavioral consequences.

Our results connect the three components. We identify controllable, robust features for all five traits (\personTerm); feature steering produces bidirectional tendencies across situational tasks while maintaining valid responses (\situationTerm) and yields trait-characteristic cost-benefit tradeoffs across social abilities (\behaviorTerm).

Our contributions are:
\textbf{(1)} We adapt the personality triad framework to connect \personTerm-side internal representations, \situationTerm-dependent responses, and broader \behaviorTerm{} outcomes in LLMs.
\textbf{(2)} We develop an SAE-based pipeline that performs feature retrieval from behavioral contrasts under matched situations, selects and causally validates candidates through intervention probing and representation analysis, and traces the selected features across personality expression and social behavior.
\textbf{(3)} We provide evidence that LLMs contain controllable trait-like representations linking internal states, situational expression, and broader behavioral outcomes.

\section{The \personTerm--\situationTerm--\behaviorTerm{} Framework}
\label{sec:framework}

\subsection{The Personality Triad Framework in Human Personality Theory}

Classical trait theory treats traits as dispositions expressed through patterned responses across situations. \citet{allport1937personality} argues that different situations can activate a common trait-related tendency while eliciting responses that differ in words, emotional tones, decisions, or actions. This shared trait relevance across situations is referred to as functional equivalence. Agreeableness, for example, may be expressed through emotional reassurance in one situation and cooperative compromise in another. These responses differ in surface form while sharing a prosocial function associated with the same trait.

\citet{FUNDER200621} extends the person--situation interaction perspective through the personality triad framework, which treats persons, situations, and behaviors as three mutually informative components. Each component is understood through its relations with the other two. A person is characterized through recurring patterns of response across situations relevant to the trait. A situation is characterized by the contextual cues and response opportunities it provides, as well as the response tendencies it elicits. A behavior acquires psychological meaning within the person--situation combination in which it occurs. The same observable action can express different functions across contexts, while different actions can instantiate a related trait tendency.

\subsection{Adapting the Personality Triad Framework for LLM Analysis}

Guided by this framework, we define three corresponding components for analyzing trait-related representations and behavior in LLMs. \personTerm{} refers to an internal representation that is activated by trait-relevant expressions and whose intervention induces trait-aligned changes while preserving coherent, instruction-following responses. \situationTerm{} refers to a concrete scenario in which the model expresses a trait-relevant tendency through a coherent, instruction-following response. \behaviorTerm{} refers to observable behavioral manifestations, reflected in performance changes across broader social tasks beyond direct personality assessment.

Our experiments examine all these components. In \rqOne, we retrieve candidate \personTerm \ representations by contrasting high- and low-trait responses within matched situations and aggregating these contrasts across diverse situations. In \rqTwo, we intervene on the selected representations and examine whether they elicit trait-relevant tendencies in both directions across a separate, diverse set of situations while preserving coherent, instruction-following responses. In \rqThree, we apply the same interventions to broader social tasks and measure the resulting changes in \behaviorTerm. Figure~\ref{fig:overview} summarizes our design.

\section{Experiments} \label{sec:experiments}\label{sec:setup}

\subsection{Experimental Setup}

\paragraph{Research scope and model.}

We use the Big Five as an established dimensional trait framework because its five broad traits (Agreeableness, Conscientiousness, Extraversion, Neuroticism, and Openness) have well-characterized high- and low-trait expressions and extensive psychological evidence concerning their behavioral correlates \cite{goldberg1990alternative,costa2008revised}. To construct fine-grained behavioral contrasts, we adopt the NEO-PI-R taxonomy of six facets within each trait \cite{costa2008revised}.

We study DeepSeek-R1-Distill-Llama-8B \cite{deepseekai2025deepseekr1incentivizingreasoningcapability} with the Llama-Scope-R1-Distill SAE \cite{he2024llamascopeextractingmillions}. This combination offers three practical advantages. First, the SAE covers residual-stream locations throughout the model, enabling a global search for personality-related features across layers. Second, the instruction-tuned model has sufficiently rich open-ended expression for intervention probing and situational advice tasks. Third, per-feature maximum activations are publicly available through Neuronpedia \cite{neuronpedia}, permitting direct and reproducible scaling of decoder directions.

\paragraph{Overall experimental design.} We organize the experiments around the three research questions. For \personTerm, contrastive behavior pairs grounded in shared situations are used for feature retrieval of sparse SAE features associated with opposing trait expressions. Intervention probing selects candidates with coherent, trait-aligned causal effects, while token-level activation and paraphrase analyses test whether the selected features generalize across lexicosyntactic forms (\rqOne). For \situationTerm, interventions on the selected features are evaluated on TRAIT \cite{lee-etal-2025-llms} for bidirectional trait expression and response validity across a separate, diverse set of situations (\rqTwo). For \behaviorTerm, the interventions are applied to SocialEval \cite{zhou-etal-2025-socialeval} to measure broader changes across interpersonal abilities (\rqThree).

\subsection{\rqOne--\personTerm: Identifying Personality- Related Internal Representations}
\label{sec:rq1}
\label{subsec:res_rq1}

\rqOne\ asks whether personality-related internal representations can be identified from contrasting behavioral expressions under matched situations, beyond lexicosyntactic cues.

We define a feature as one dimension of the SAE latent representation, or equivalently the hidden-state direction obtained by decoding that dimension. The pretrained SAE provides a sparse decomposition of model activations, allowing us to retrieve candidate features whose activation patterns distinguish the two response poles using a controlled dataset. During intervention, each decoded direction is added at the residual-stream position of its corresponding SAE, ensuring consistency between feature retrieval and causal manipulation. Dataset construction, intervention probing and validation protocols, and supplementary representation analyses are provided in Appx.~\ref{appendix:rq1_details}.

\begin{table}[htbp]
\centering
{
\scriptsize
\setlength{\tabcolsep}{1mm}
\begin{tabularx}{\columnwidth}{X X X X}
\toprule
\textbf{Trait} & \textbf{(L, Idx)} & \textbf{High/Low/{$\Delta f$} (Original)} & \textbf{High/Low/{$\Delta f$} (Paraphrased)} \\ \midrule
Agreeableness      & (9, 525)    & 202 / 45 / 157  & 224 / 64 / 160  \\
Conscientiousness  & (7, 8233)   & 347 / 31 / 316  & 349 / 20 / 329  \\
Extraversion       & (13, 27392) & 429 / 110 / 319 & 446 / 148 / 298 \\
Neuroticism        & (12, 22254) & 322 / 48 / 274  & 252 / 101 / 151 \\
Openness           & (6, 4344)   & 234 / 102 / 132 & 208 / 113 / 95  \\ \bottomrule
\end{tabularx}
}
\caption{Activation statistics for the selected features on the original and paraphrased matched-situation pairs.}
\label{tab:paraphrased_stats}
\end{table}

\subsubsection{Discovering Personality-Related Features}

\paragraph{Contrastive behavior-pair construction.} Let {$\mathcal{C}=\{s_1,\ldots,s_K\}$} be a situational corpus and {$\mathcal{T}$} the target traits. We instantiate {$\mathcal{C}$} with Q-Sort situational corpus \cite{Neuman2023} extended from the Riverside Situational Q-Sort \cite{doi:10.1177/0963721416635552}, and use NEO-PI-R traits and facets to define opposing behavioral tendencies \cite{costa2008revised}. For each {$t\in\mathcal{T}$}, LLM filtering and expert audit define a validated subset {$\mathcal{C}_t=\Phi_t(\mathcal{C})$}, where each retained situation is mapped to one facet. For each {$s\in\mathcal{C}_t$}, an LLM-based expansion function generates concise high- and low-facet reactions that share the same situation clause:
{
\begin{equation}
\mathcal{X}_s = \Psi(s, t; \mathcal{P}_t) = \{ (x_k^+, x_k^-) \}_{k=1}^{K_s}
\end{equation}
}%
The per-trait Feature Retrieval Dataset and Intervention Probing Dataset are
{
\small
\begin{equation}
\begin{aligned}
\mathcal{D}_{\text{ret}}^{(t)}
&=\operatorname{UnifSample}_{\mathrm{facet}}\!\left(\bigcup_{s\in\mathcal{C}_t}\mathcal{X}_s\right),\\
\mathcal{D}_{\text{probe}}^{(t)}
&=\{(s,Q(s))\mid s\in\operatorname{UnifSample}_{\mathrm{facet}}(\mathcal{C}_t)\}.
\end{aligned}
\end{equation}
}%
where {$K_s$} is the number of contrastive pairs generated for {$s$}, and {$Q(\cdot)$} instantiates an open-ended probing question. Pairs in {$\mathcal{D}_{\mathrm{ret}}^{(t)}$} are evenly sampled across the six facets of each trait, yielding 500 pairs per trait. Contrasting high- and low-trait behaviors within each of diverse situations reflects the conception of a trait as a stable tendency toward functionally equivalent responses across situations \cite{allport1937personality}. Dataset construction details and examples are provided in Appx.~\ref{appendix:datasetconstructionandinstantiation} and Appx.~\ref{appendix:datasetexamples}.

\paragraph{SAE encoding.} For token {$k$}, the model hidden state {$\mathbf{h}_k$} is mapped to SAE activations {$\mathbf{f}_k$}. To obtain a sequence-level representation while preserving features that respond strongly at specific token positions, we max-pool across the sequence:
{
\begin{equation}
\mathbf{F} = \text{max\_pool}(\mathbf{f}_1, \mathbf{f}_2, \dots, \mathbf{f}_T)
\end{equation}
}%
Under SAE sparsity, a target feature should activate frequently on one pole and remain suppressed on the other. Because each positive--negative pair shares trait-unrelated lexicosyntactic forms and semantics, unrelated features should have similar activation frequencies in the two sets. For feature {$i$}, we first calculate its activation count and rate on each pole:
{
\begin{equation}
\begin{aligned}
N_i^{\mathrm{pol}} &= \sum_{j=1}^{500}\mathbb{I}(F_{i,j}^{\mathrm{pol}}>0),\\
P_i^{\mathrm{pol}} &= \frac{N_i^{\mathrm{pol}}}{500},
\quad \mathrm{pol}\in\{\mathrm{pos},\mathrm{neg}\}.
\end{aligned}
\end{equation}
}%
We retain features with a sufficiently large frequency difference and a nontrivial activation rate on at least one pole:
{
\begin{equation}
\scalebox{0.85}{$\displaystyle
\mathcal{S} = \{\, i \,\mid\, |N_i^{\mathrm{pos}}-N_i^{\mathrm{neg}}| \geq \tau_1 \;\land\; \max(P_{i}^{\text{pos}},\, P_{i}^{\text{neg}}) \geq \tau_2 \,\}
$}
\end{equation}
}%
where {$\mathcal{S}$} is the retained candidate set. We set {$\tau_1=80$} and {$\tau_2=0.2$}, fixed to retain features that distinguish the two poles across multiple facets. This stage identifies activation-associated candidates, which subsequently undergo intervention probing because similar input-side statistics can yield substantially different steering effects (Appx.~\ref{appendix:discrepancybetweeninputactivationsandsteeringeffectiveness}).

\paragraph{Feature intervention.} For candidate feature {$i$} at layer {$l_i$}, we reconstruct its decoder direction and define
{
\begin{equation}
\mathbf{v}_{\text{steer}} = \alpha \cdot \phi_{i} \cdot \mathbf{W}_{\text{dec}}^{(i)}
\end{equation}
\begin{equation}
\mathbf{h}_l' = \mathbf{h}_l + \mathbf{v}_{\text{steer}}
\end{equation}
}%
where {$\alpha$} is the steering coefficient, {$\phi_i$} is the feature's maximum activation during SAE training, and {$\mathbf{W}_{\mathrm{dec}}^{(i)}\in\mathbb{R}^d$} is the {$i$}-th column of the SAE decoder matrix, with {$d$} denoting the hidden dimension. Following prior SAE steering practice \cite{templeton2024scaling}, scaling by {$\phi_i$} keeps the intervention commensurate with activation magnitudes observed during training.

\paragraph{Generation-based intervention probing.} For each candidate {$i\in\mathcal{S}$}, we sweep {$\mathcal{A}=\{0,\pm0.25,\pm0.5,\pm1\}$} and collect an ordered response family {$\{y_q^{(i,\alpha)}\}_{\alpha\in\mathcal{A}}$} for every {$q\in\mathcal{D}_{\text{probe}}^{(t)}$}. Given the trait definition, descriptions of its high and low behaviors, and the ordered responses, the hybrid judge {$\mathcal{J}$} combines an initial LLM assessment with expert audit and returns
{
\begin{equation}
c_q^{(i)} = \mathcal{J}\!\bigl(t,\, q,\, \{y_q^{(i,\alpha)}\}_{\alpha \in \mathcal{A}}\bigr) \in \{0,1\},
\end{equation}
}%
where {$c_q^{(i)}=1$} when the responses are grammatical and coherent and show a clear polarity change consistent with the trait's high--low behaviors as {$\alpha$} varies. Because feature orientation is arbitrary, the judge accepts either monotonic direction and rejects uncertain or invalid cases. A candidate feature is selected if it receives at least one positive label across its associated questions, i.e., {$\sum_{q\in\mathcal{D}_{\text{probe}}^{(t)}}c_q^{(i)}>0$}. The prompts and audit procedure are given in Appx.~\ref{appendix:featurevalidationprotocol}; an judge-reliability experiment is provided in Appx.~\ref{appendix:reliabilityofthellmjudge}.

\subsubsection{Characterizing Personality-Related Features}

\begin{figure}[htbp]
    \centering
    \includegraphics[width=\linewidth]{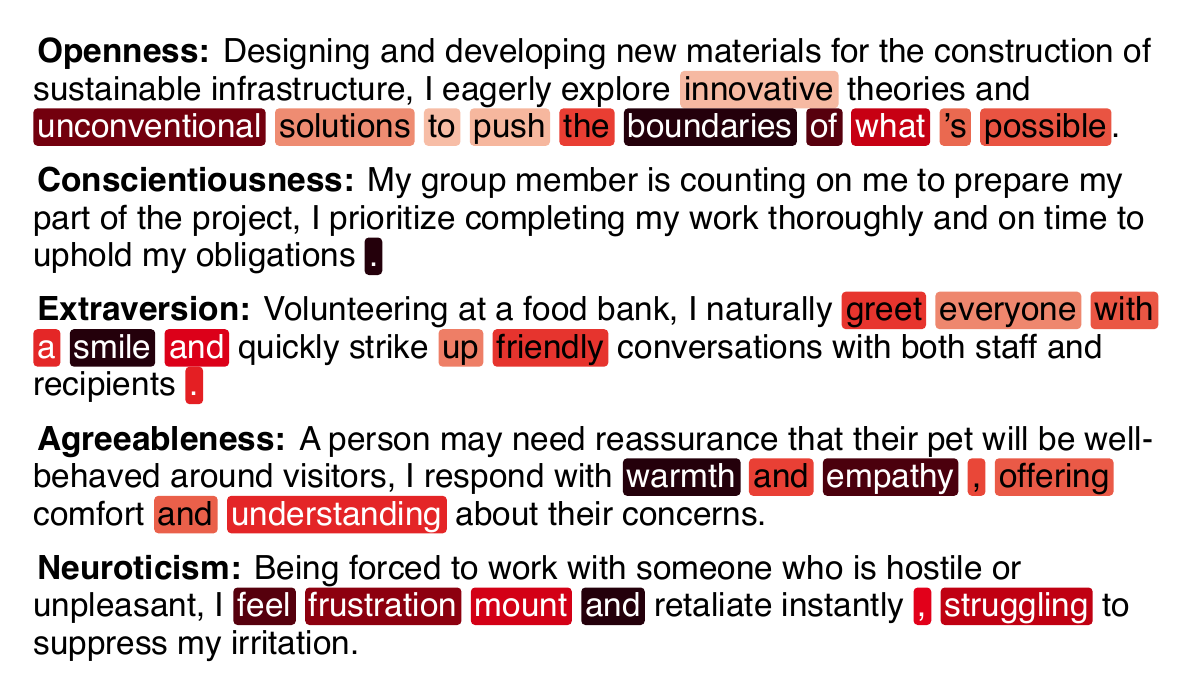}
    \caption{Token-level activations of the selected SAE features. Darker highlights indicate higher activation values.}
    \label{fig:heatmap}
\end{figure}

\paragraph{Token-level activation.} We inspect where the selected features activate in the Feature Retrieval Dataset. Figure~\ref{fig:heatmap} shows two recurring patterns. Activations may be distributed over trait-relevant words or phrases, consistent with the semantic content of the trait; they may also peak at sequence boundaries, indicating sensitivity to the trait-related meaning of the preceding clause as a whole. For example, Agreeableness is associated with prosocial expressions such as empathy and compassion, whereas the selected Conscientiousness feature often peaks near sequence boundaries. Quantitative token statistics are present in Appx.~\ref{appendix:quantitativeanalysisoftokenactivationcorrelation}.

\paragraph{Robustness to paraphrase.} A personality-related representation should capture trait meaning rather than specific lexicosyntactic forms. We therefore rewrite the Feature Retrieval Dataset, substantially changing vocabulary and syntax while preserving each situation and high--low facet contrast. The selected features are then evaluated with the same thresholds. Table~\ref{tab:paraphrased_stats} shows that their activation-frequency differences remain above {$\tau_1$}. For example, Conscientiousness has a difference of 329 and a maximum activation frequency of 349 out of 500 pairs after paraphrasing, compared with a difference of 316 on the original set. These results indicate that the selected features capture trait-related meaning across substantially different lexicosyntactic forms. Prompts and examples are provided in Appx.~\ref{appendix:paraphraserobustnessprotocol}.

Together, feature discovery identifies personality -related features from contrasting behaviors within shared situations, while feature characterization further validates their trait relevance and semantic grounding.

\begin{table*}[t]
\centering
{
\scriptsize
\setlength{\tabcolsep}{1mm}
\begin{tabularx}{\textwidth}{X X X X X X}
\toprule
\textbf{Trait} & \textbf{Method} & \textbf{(L, Idx)} & \textbf{Polarity} & \textbf{Trait Score} & \textbf{Valid Rate} \\ \midrule
\multirow{4}{*}{Agreeableness}
    & Baseline & - & - & 0.7052 & 0.960 \\
    & CAA & (9, -) & {$\pm$} & 0.8070 / 0.4220 & 0.993 / 0.987 \\
    & {$P^2$} & - & {$\pm$} & 0.7583 / 0.6095 & 0.989 / 0.968 \\
    & Ours & (9, 525) & {$\pm$} & 0.7845 / 0.6278 & 0.942 / 0.994 \\ \midrule
\multirow{4}{*}{Conscientiousness}
    & Baseline & - & - & 0.8695 & 0.958 \\
    & CAA & (7, -) & {$\pm$} & 0.7150 / 0.7630 & 0.930 / 0.800 \\
    & {$P^2$} & - & {$\pm$} & 0.8609 / 0.8217 & 0.985 / 0.976 \\
    & Ours & (7, 8233) & {$\pm$} & 0.9043 / 0.8294 & 0.961 / 0.985 \\ \midrule
\multirow{4}{*}{Extraversion}
    & Baseline & - & - & 0.4463 & 0.977 \\
    & CAA & (13, -) & {$\pm$} & 0.2950 / 0.1430 & 0.924 / 0.862 \\
    & {$P^2$} & - & {$\pm$} & 0.5724 / 0.2146 & 0.987 / 0.983 \\
    & Ours & (13, 27392) & {$\pm$} & 0.6609 / 0.3971 & 0.985 / 0.972 \\ \midrule
\multirow{4}{*}{Neuroticism}
    & Baseline & - & - & 0.2117 & 0.959 \\
    & CAA & (12, -) & {$\pm$} & 0.9290 / 0.1270 & 0.141 / 0.283 \\
    & {$P^2$} & - & {$\pm$} & 0.2208 / 0.1261 & 0.969 / 0.983 \\
    & Ours & (12, 22254) & {$\pm$} & 0.4412 / 0.1017 & 0.961 / 0.944 \\ \midrule
\multirow{4}{*}{Openness}
    & Baseline & - & - & 0.5214 & 0.959 \\
    & CAA & (6, -) & {$\pm$} & 0.6140 / 0.3520 & 0.938 / 0.971 \\
    & {$P^2$} & - & {$\pm$} & 0.6161 / 0.4030 & 0.969 / 0.990 \\
    & Ours & (6, 4344) & {$\pm$} & 0.5436 / 0.5164 & 0.951 / 0.947 \\
\bottomrule
\end{tabularx}
}
\caption{TRAIT situational-response results. For intervention methods, statistics are reported for both polarity.}
\label{tab:personality_results}
\end{table*}

\subsection{\rqTwo--\situationTerm: Personality Expression across Situations}
\label{sec:rq2}
\label{subsec:res_rq2}

\rqTwo\ asks whether intervention on the identified representations elicits bidirectional trait-relevant tendencies across a separate, diverse set of situations while preserving coherent, instruction-following responses.

\subsubsection{Situational-Response Evaluation}

We use the psychometrically validated Big Five subset of TRAIT \cite{lee-etal-2025-llms}. Each item presents a concrete, realistic user situation and asks the model for advice, with available choices reflecting different trait tendencies. The official next-token probability protocol measures statistical preference over these choices but cannot reveal how the intervened model actually responds to the situation. We therefore extend it with open generation, requiring the model to produce reasoning and explicitly select an option. An automated extractor identifies the choice and marks incoherent, instruction-violating, or unextractable responses as invalid.

We report \textit{trait score}, the proportion of high-trait selections among valid responses, and \textit{valid rate}, the proportion of responses not marked invalid. Trait score summarizes directional expression over the benchmark's collection of situations; valid rate establishes whether the model continues to produce coherent, instruction-following responses under intervention. The two metrics jointly characterize successful situated expression. A trait-score shift alone does not establish that the model continues to engage with the situation, because severe intervention may leave only a small subset of responses coherent and extractable. Conversely, a high valid rate without directional change indicates preserved generation but ineffective regulation. Successful intervention therefore requires both a consistent trait-directional shift and continued production of coherent, instruction-following responses.

\subsubsection{Comparative Interventions}

The no-intervention model provides the original trait tendency. We additionally evaluate two controls from different intervention levels: {$P^2$} \cite{NEURIPS2023_21f7b745}, which assigns the target personality through a persona prompt at the input level, and CAA \cite{rimsky-etal-2024-steering}, which injects a contrastive mean-difference direction into the residual stream. The selected SAE-feature intervention uses {$\alpha=\pm1$}; for CAA, we use {$\alpha=\pm2$}, following the setting demonstrated in the original study. The comparisons characterize each concrete intervention in terms of bidirectional regulation and preservation of coherent, instruction-following responses.

\subsubsection{Results: Expression across Situations}

\paragraph{Bidirectional regulation.} As shown in Table~\ref{tab:personality_results}, the selected feature intervention produces the expected positive--baseline--negative ordering for all five traits. The clearest shifts occur for Extraversion (0.661/0.397 around a 0.446 baseline) and Neuroticism (0.441/0.102 around 0.212). Across traits, the same \personTerm-side representation regulates trait-relevant choices over a separate, diverse set of situations.

\paragraph{Preserved situated response.} Valid rates remain close to the no-intervention model throughout, including 0.961/0.944 for Neuroticism compared with 0.959 at baseline and 0.961/0.985 for Conscientiousness compared with 0.958. The intervention changes personality-related choices while preserving the model's ability to understand the scenario, produce coherent advice, and complete the required selection.

\paragraph{Intervention effectiveness and response validity.} The controls show that both bidirectional trait change and response validity are important for evaluating an intervention. For Conscientiousness, the nominal positive condition of {$P^2$} moves the baseline from 0.870 to 0.861, and its negative condition reaches 0.822; for Neuroticism, its positive condition changes 0.212 to 0.221. This particular persona intervention therefore provides limited bidirectional regulation for these traits. CAA can produce large score changes, but on Neuroticism it reduces validity to 0.141/0.283, leaving few responses that reveal how the model addresses the situation. Its validity also falls to 0.800 in negative Conscientiousness and to 0.862 in negative Extraversion. These failures show that an internal intervention can disrupt the coherent, instruction-following situated response on which personality observation depends (examples in Appx.~\ref{appendix:failedcasesofcaa}).

These results answer \rqTwo: the \personTerm-side representations identified in \rqOne{} elicit trait-relevant tendencies in both directions across a separate, diverse set of situations while preserving coherent, instruction-following responses.

\subsection{\rqThree--\behaviorTerm: Psychologically Consistent Behavioral Changes}
\label{sec:rq3}
\label{subsec:res_rq3}

\rqThree\ asks whether the same intervention induces consistent changes across social behaviors that correspond to findings in human personality research.

\subsubsection{Behavioral Evaluation with SocialEval}

TRAIT directly probes trait-relevant situational choices. To measure trait-related behavior in broader social scenarios, we use the SocialEval Interpersonal Ability Evaluation (IAE) \cite{zhou-etal-2025-socialeval}, which evaluates interpersonal abilities such as teamwork and emotional regulation through heterogeneous social scripts. We apply the same selected features with {$\alpha=\pm1$} and report accuracy for each interpersonal ability.

\begin{figure}[ht]
    \centering
    \includegraphics[width=1\linewidth]{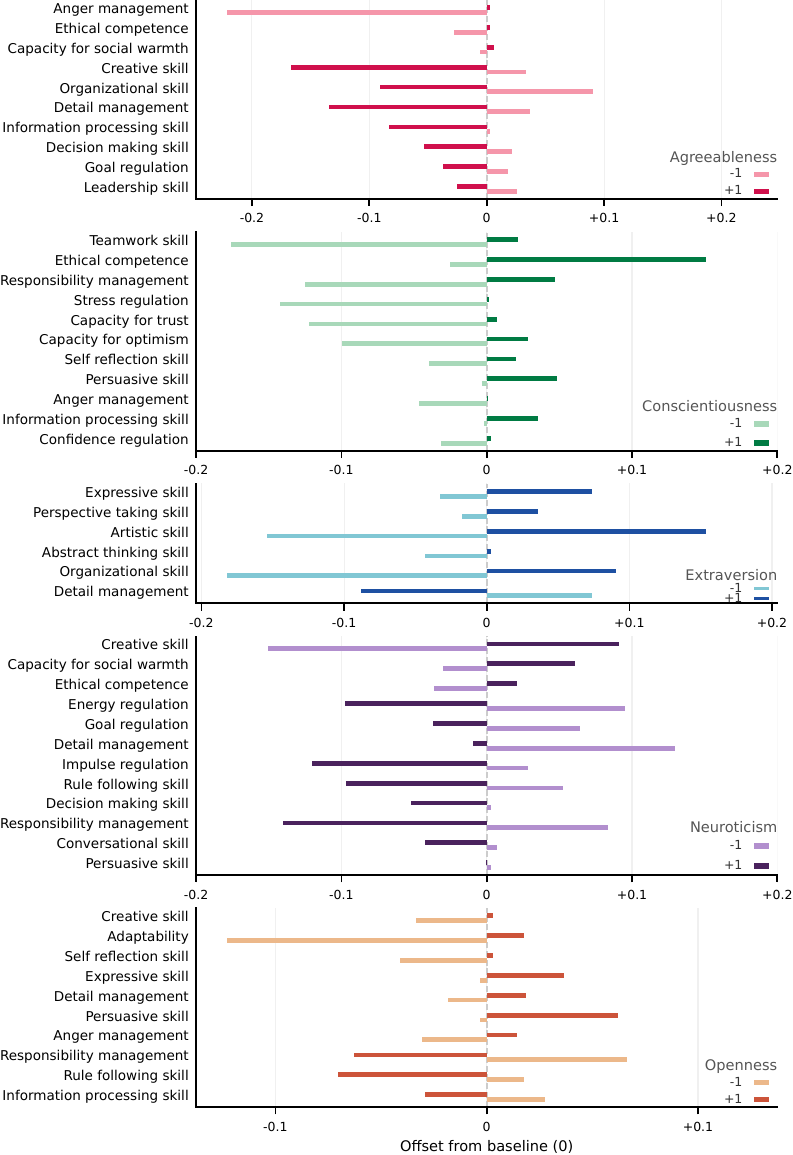}
    \caption{Representative SocialEval changes under personality-related feature intervention. Positive and negative shifts form trait-characteristic benefit--cost patterns consistent with meta-analytic findings in personality psychology.}
    \label{fig:socialeval_iae}
\end{figure}

\subsubsection{Behavioral Results}

Figure \ref{fig:socialeval_iae} shows systematic, trait-specific benefit-cost patterns consistent with meta-analytic findings in personality psychology. \cite{barrick1991big,habashi2016searching,pletzer2019meta,costa2008revised}. Specifically, \textit{Agreeableness} strengthens prosocial and conflict-regulation abilities relative to negative steering, including anger management and ethical competence, while reducing performance on several self-agency and execution-oriented tasks \cite{wilmot2022agreeableness}. \textit{Conscientiousness} produces gains in responsibility management, ethical competence, teamwork, and regulation-related abilities, corresponding to its established associations with responsibility, self-regulation, and goal-directed behavior \cite{roberts2009conscientiousness,jackson2010conscientious,eisenberg2014conscientiousness}. \textit{Extraversion} improves social-expression and interaction-related performance, including expressive skill and perspective taking, with additional gains in artistic and organizational skills, while reducing performance on tasks requiring sustained focus or fine control, such as detail management \cite{john2008paradigm,deyoung2007between,fishman2011extraverts}. Conversely, \textit{Neuroticism} weakens regulatory and executive-control abilities, including goal regulation and rule following, while improving performance on creative skill and capacity for social warmth \cite{watson1984negative,lahey2009public}. \textit{Openness} improves adaptability, expressive skill, and persuasive skill, and yields higher creative and self-reflective performance under positive than negative steering, while reducing performance on structured tasks such as responsibility management \cite{deyoung2015cybernetic,mccrae1987creativity}.

Overall, the changes in the model's behavioral patterns correspond to established experimental findings in personality psychology. Rather than producing uniform improvement or degradation, each intervention yields a differentiated profile of benefits and costs across heterogeneous social tasks. These
findings demonstrate psychologically consistent changes in
broader social behaviors, answering \rqThree. Detailed per-ability scores and trait-wise discussion are provided in Appx.~\ref{appendix:moreanalysisofsocialevalresults}.

\section{Discussion and Conclusion} \label{sec:discussion}\label{sec:conclusion}

This work adapts the personality triad framework to study personality-related representations and behavior in LLMs. The three research questions establish a connected chain of evidence across internal activations, trait scores, and behavioral outcomes. In \rqOne, we retrieve candidate SAE features from contrasting behaviors under matched situations and select those whose intervention causally changes trait-relevant generation; token-level and paraphrase analyses further establish their semantic grounding beyond lexicosyntactic forms. In \rqTwo, the same features regulate trait-related expression bidirectionally across a separate, diverse set of situations while preserving coherent, instruction-following responses. Analysis of alternative interventions further shows that observing the representation--expression relation requires both changing the intended tendency and preserving coherent, instruction-following responses. In \rqThree, applying the same interventions to heterogeneous social tasks produces characteristic combinations of benefits and costs that correspond to findings in human personality research. Together, these results provide evidence that LLMs contain controllable trait-like representations that connect \personTerm-side internal states, \situationTerm-dependent expression, and broader \behaviorTerm{} outcomes.

\section*{Limitations}
Several scope choices bound the conclusions of this study. All stimuli, prompts, and benchmarks were developed in English; consequently, cross-linguistic generalizability needs further examination. Furthermore, we adopted the Big Five as the primary trait framework, utilizing the NEO-PI-R for facet-level structure. As alternative dimensional systems were not investigated, the findings may not extend to other personality models.

{
\bibliography{custom}
}

\appendix

\section{Related Work} \label{sec:related_work}

\subsection{Personality Induction and Evaluation in LLMs}

Personality induction aims to endow an LLM with a target profile so that its responses change accordingly across interactions. Existing methods span persona prompting, weight-space editing, training-based alignment, and mechanistic intervention on internal representations \cite{brito2025modeling,ye2026largelanguagemodelpsychometrics}. Prompting approaches place a persona description or trait instruction in the input \cite{chen2024from,Serapio-Garcia2025,pan2023llmspossesspersonalitymaking}. Weight-space approaches alter or compose parameters, including model-merging personality vectors \cite{sun-etal-2025-personality} and trait-specific adapters \cite{Vu2026}, while training-based approaches use supervised or preference learning over trait-annotated data \cite{li-etal-2025-big5}.

Evaluation methods can be organized by their form of elicitation. Inventory-based assessment asks models to answer human personality scales, either directly or under target-profile conditioning. Existing studies examine the applicability of self-assessment questionnaires \cite{song2023largelanguagemodelsdeveloped}, construct a psychometric framework around Big Five inventories \cite{Serapio-Garcia2025}, use Big Five questionnaires to evaluate personality emulation \cite{wang2025evaluating}, or use a modified MBTI questionnaire to evaluate personality-trained models \cite{cui2024machinemindsetmbtiexploration}. Situated-choice assessment transforms psychometric constructs into multiple-choice decisions grounded in real-world scenarios, as in TRAIT \cite{lee-etal-2025-llms}. Generation-based assessment elicits open-ended answers and maps their linguistic expression to numerical trait estimates, as in LMLPA \cite{10.1162/coli_a_00550}. These elicitation forms use different scoring interfaces: direct inventories record a closed-ended, single-option response for each item, TRAIT compares option-token probabilities, and LMLPA applies an automated rater to generated text.

Some work reports stable or distinct self-report personality profiles \cite{huang-etal-2024-reliability,heston2025large}, while other studies identify important measurement problems: self-assessment tests can be unreliable measures of LLM personality \cite{gupta-etal-2024-self}, response statistics can deviate from human patterns \cite{10.1145/3757887.3763016}, and personality estimates can vary across scales \cite{tosato2024llms}. More directly, persona injection may shift self-reported scores while leaving situational behavior only weakly aligned \cite{han2025personality}. These findings motivate us to connect measured tendencies with meaningful responses and their behavioral consequences. Our study therefore traces an identified internal representation through situational expression and into social behavior.

\subsection{Mechanistic Interpretability of Personality-related Representations in LLMs}

Mechanistic interpretability offers a route from correlation to control for LLM personality by localizing behaviorally relevant representations inside a model \cite{bdcc9080193}. This approach is often motivated by the linear representation hypothesis that semantic and behavioral attributes occupy linearly separable subspaces \cite{pmlr-v235-park24c}. Contrastive Activation Addition (CAA) retrieves a direction as the mean activation difference over contrastive stimuli and adds it during inference \cite{rimsky-etal-2024-steering}. Persona Vectors similarly identify activation-space directions for monitoring and controlling character traits \cite{chen2025personavectorsmonitoringcontrolling}. Neuron- and head-level methods localize trait-related units at finer granularity, including NPTI \cite{ICLR2025_d399b67f} and PAS \cite{ICLR2025_25203d1c}.

Two issues are especially relevant when these tools are used within the personality triad framework. First, contrastive retrieval can retain lexicosyntactic patterns from its probing data, limiting representational abstraction. Second, a dense mean-difference vector can be sensitive to probing noise and may impair generation at large intervention strengths \cite{rimsky-etal-2024-steering}, thereby removing the model's ability to respond effectively to a situation. Sparse autoencoders (SAEs) offer a complementary route by decomposing dense hidden states into sparse, more monosemantic features \cite{bricken2023monosemanticity,templeton2024scaling,lieberum-etal-2024-gemma,he2024llamascopeextractingmillions}. Such features have been used to localize linguistic phenomena \cite{jing-etal-2025-lingualens} and study behaviors such as repetition \cite{yao-etal-2025-understanding}. We use SAE features as candidate \personTerm-side representations, test their generalization across lexicosyntactic forms, and use intervention to trace their effects across \situationTerm{} and \behaviorTerm.

\section{\rqOne--\personTerm: Feature Retrieval, Intervention Probing, and Validation}
\label{appendix:rq1_details}

\subsection{Dataset Construction and Instantiation}
\label{appendix:datasetconstructionandinstantiation}
\raggedbottom The construction of our dataset follows a rigorous pipeline where Qwen3-235B-Thinking \cite{yang2025qwen3technicalreport} serves as the primary engine for situation filtering and Feature Retrieval Dataset generation. First, 100 situational categories from the Q-Sort \cite{Neuman2023} dataset are processed; for each category, the model identifies personality facets that can be sufficiently manifested within that context, if possible. The initial retrieval results undergo review and labeling to ensure each selected situation is mapped to a unique trait-facet pair. Specifically, three experts independently label each situation with the most suitable facet it demonstrates, and we adopt a majority vote (at least 2/3 agreement) to retain a situation. Subsequently, for each refined situational category, the model is tasked to expand it into specific scenarios and generate contrastive reaction pairs representing high and low scores on the targeted facet. The filtering results by LLM and the subsequent inter-rater agreement among experts are summarized in Table~\ref{tab:situation_stats} and Table~\ref{tab:situation_expert_agreement} respectively. Finally, contrastive pairs are evenly sampled among the six facets of each trait, with uniform sampling applied per situation, yielding 500 pairs per trait (2,500 total). For the Intervention Probing Dataset, we randomly select one valid situation per facet and append an open-ended prompt to elicit trait-relevant generation. This yields 30 probing questions in total, with each candidate feature evaluated against the six questions corresponding to its associated trait's facets.

\begin{table}[htbp]
\centering
{
\scriptsize
\setlength{\tabcolsep}{1mm}
\begin{tabularx}{\columnwidth}{    l@{\hspace{18mm}}
    c}
\hline
\textbf{Personality Trait} & \textbf{Number of Situations} \\ \hline
Extraversion & 33 \\
Agreeableness & 38 \\
Conscientiousness & 50 \\
Neuroticism & 38 \\
Openness & 19 \\ \hline
\end{tabularx}
}
\caption{Number of Situations Suitable for Each Trait Filtered by LLM.}
\label{tab:situation_stats}
\end{table}

\begin{table}[htbp]
\centering
{
\scriptsize
\setlength{\tabcolsep}{1mm}
\begin{tabularx}{\columnwidth}{    l@{\hspace{22mm}}
    c}
\hline
\textbf{Trait} & \textbf{Fleiss' Kappa} \\ \hline
Extraversion & 0.8207 \\
Agreeableness & 0.6039 \\
Conscientiousness & 0.6149 \\
Neuroticism & 0.7759 \\
Openness & 0.9334 \\ \hline
\end{tabularx}
}
\caption{Inter-rater Agreement Statistics (Fleiss' Kappa, {$n=3$}) for Expert Facet Labeling.}
\label{tab:situation_expert_agreement}
\end{table}

\begin{examplebox}{Task Guidelines for Expert Labeling}
\textbf{1. Project Overview} \\
This task aims to validate the psychological relevance of various ``Situations'' designed to elicit distinct behaviors from individuals with high or low scores in specific personality traits. As a psychology expert, your goal is to identify which specific \textbf{Facet} of a given \textbf{Trait} is most effectively demonstrated by the provided situation.

\textbf{2. Operational Protocols} \\
This is an \textbf{Independent Expert Review} task. Please adhere to the following phases:
\begin{itemize}
    \item \textbf{Phase I: Contextual Analysis.} Review the provided Trait and its six Facets. A situation is well-matched if it naturally forces a choice that distinguishes a ``High Scorer'' from a ``Low Scorer''.
    \item \textbf{Phase II: Independent Labeling.} \textit{Forced Choice:} Select the single facet that best illustrates the situation based on \textbf{Relevance} and \textbf{Discriminative Power}. Use ``None of the above'' only if completely irrelevant.
    \item \textbf{Phase III: Justification.} Provide a concise, one-sentence psychological rationale (e.g., \textit{``The scenario involves a direct threat to social standing...''}).
\end{itemize}

\textbf{Note:} Do not consult with other experts during this process. Your independent professional judgment is the primary data point.
\end{examplebox}

\begin{examplebox}{System Prompt of Situation Category Annotation}
\label{prompt1}
\begin{Verbatim}[fontsize=\small,breaklines=true, breakanywhere=true, commandchars=\\\{\}]
# Task Instructions

## Your Role
You are an expert annotator specializing in personality psychology. Your task is to analyze and annotate various situations based on established psychological theories and frameworks.

## Your Task
1. **Situation Analysis**: Carefully read and understand the provided situation, which includes multiple examples illustrating the context.
2. **Annotation**: Based on your analysis, provide a concise annotation that captures the essence of the situation. Then, analysis which trait(s) from the Big Five personality traits (Openness, Conscientiousness, Extraversion, Agreeableness, Neuroticism) are most relevant to the situation. Justify your choice with a brief explanation. Your annotation should be clear, informative, and relevant to personality psychology.

## Input Format
You will receive input in the following format:

```
# Situation Name
<name of the situation>

# Situation Examples
<example 1>
<example 2>
...
<example n>
```

## Output Format
Your response should be structured in the following JSON format:

```json
{
  "annotation": "Your concise annotation of the situation.",
  "related_traits": [
    {
      "trait": "Name of the related Big Five trait",
      "justification": "Brief explanation of why this trait is relevant to the situation."
    },
    {... Additional traits if applicable ...}
  ]
}
```

## Important Notes
- Ensure that your annotations are based on established psychological theories and frameworks.
- Be objective and avoid personal biases in your analysis.
- If the situation does not clearly relate to any of the Big Five traits, you may indicate that no traits are applicable, and return an empty list for "related_traits". If there are multiple relevant traits, include all applicable ones with justifications.

## Additional Information

- Five-trait mnemonics:
  - Conscientiousness: self-discipline, planning, rule-following (positively linked to achievement and health).
  - Agreeableness: cooperation, compassion, harmony-seeking (positively linked to prosocial behavior; may increase obedience to authority).
  - Extraversion: sociability, energy, reward-seeking in social contexts (linked to social interaction and leadership).
  - Openness: curiosity, creativity, novelty-seeking (linked to innovation and some risk-taking).
  - Neuroticism: emotional instability, anxiety (linked to interpersonal conflict and social avoidance).
\end{Verbatim}
\end{examplebox}

\begin{examplebox}{System Prompt of Feature Retrieval Dataset Generation}
\label{prompt2}
\begin{Verbatim}[fontsize=\small,breaklines=true, breakanywhere=true]
**Role:** You are a personality psychology expert specializing in the Five-Factor Model (Big Five) and its 30 facets as described by the NEO-PI-R. Your task is to provide nuanced insights into how different personality facets might influence a person's behavior in a given scenario.

**Instructions:**
You will be provided with a **situation**, a specific **Big Five trait**, a corresponding **Big Five facet**, and a **description** of that trait and facet. Based on this information, you will write two separate sentences.
* **Sentence 1** should describe the reaction of a person who scores **high** on the specified facet and corresponding Big Five trait.
* **Sentence 2** should describe the reaction of a person who scores **low** on the specified facet and corresponding Big Five trait.
* Each sentence must consist of two clauses, with correct grammatical and semantic structure:
    * **Clause 1:** A description of the **situation**. This clause must be identical for both sentences. The situation can be directly quoted from the input or slightly rephrased, but the core meaning must remain unchanged. You should decide whether to use first-person or third-person perspective based on the situation description, so that the second clause can clearly illustrate the high or low facet trait. You should also ensure that the situation is described in a natural and coherent manner considering the second clause.
    * **Clause 2:** A description of the first-person reaction, clearly illustrating the high or low facet trait.
* Ensure the output is a list with exactly two sentences.

**Example:**
* **Situation:** A presentation to a new team tomorrow.
* **Big Five Trait:** Neuroticism
* **Big Five Trait Description:** Measures emotional stability and a person's tendency to experience negative emotions. One with high Neuroticism tends to be emotionally unstable, prone to experiencing negative emotions like anxiety, anger, and depression. One with low Neuroticism is emotionally stable, able to handle stress calmly, and rarely feels nervous or discouraged.
* **Facet:** Anxiety
* **Facet Description:** One with high Anxiety is habitually worried and tense, even when things are going well. One with low Anxiety is calm and composed, typically not bothered by small things.

**Output:**
["Facing a presentation to a new team tomorrow, I am overwhelmed with worry about potential mistakes and how I will be perceived.", "Facing a presentation to a new team tomorrow, I remain composed and confident, focusing on delivering my message effectively."]

**Note:** 
*   Each sentence should be concise.
*   You should only provide the two sentences as output without any additional commentary or explanation.
\end{Verbatim}
\end{examplebox}

\subsection{Dataset Examples}
\label{appendix:datasetexamples}

\begin{examplebox}{I. Feature Retrieval Dataset Examples}

\textbf{Case 1: Agreeableness (Tender-mindedness)}
\begin{itemize}[leftmargin=1.5em, nosep]
    \item \textbf{Situation:} After a break-up, someone may need reassurance that they will find love again.
    \item \textbf{High Reaction:} I offer warm, empathetic support, comforting them with kindness and hope.
    \item \textbf{Low Reaction:} I focus on the practical realities, preferring logic over emotional comfort.
\end{itemize}

\textbf{Case 2: Conscientiousness (Dutifulness)}
\begin{itemize}[leftmargin=1.5em, nosep]
    \item \textbf{Situation:} My boss is counting on me to finish a project by the end of the day.
    \item \textbf{High Reaction:} I meticulously organize my tasks to fulfill the deadline as agreed.
    \item \textbf{Low Reaction:} I procrastinate and dismiss the urgency of completing it on time.
\end{itemize}

\textbf{Case 3: Extraversion (Activity)}
\begin{itemize}[leftmargin=1.5em, nosep]
    \item \textbf{Situation:} Going to a karaoke night and having fun singing with friends.
    \item \textbf{High Reaction:} I energize the room by choosing upbeat songs and encouraging others to join.
    \item \textbf{Low Reaction:} I observe performances and sing a few songs at my own leisure.
\end{itemize}

\textbf{Case 4: Neuroticism (Depression)}
\begin{itemize}[leftmargin=1.5em, nosep]
    \item \textbf{Situation:} Getting stuck in a traffic jam when running late for an important meeting.
    \item \textbf{High Reaction:} I feel overwhelmed by a sense of hopelessness; nothing will ever go right.
    \item \textbf{Low Reaction:} I remain positive and focus on practical solutions without succumbing to discouragement.
\end{itemize}

\textbf{Case 5: Openness (Ideas)}
\begin{itemize}[leftmargin=1.5em, nosep]
    \item \textbf{Situation:} Developing and launching new products in the technology industry.
    \item \textbf{High Reaction:} I thrive on brainstorming novel approaches and diving into frontier concepts.
    \item \textbf{Low Reaction:} I prefer sticking to proven methods and avoid abstract or hypothetical debates.
\end{itemize}
\end{examplebox}

\begin{examplebox}{II. Intervention Probing Dataset Examples}
\textbf{Methodology:} These prompts are used to probe candidate features for coherent, trait-aligned effects after intervention across all 30 facets defined by NEO-PI-R of the Big Five model.

\begin{description}[leftmargin=1em, font=\bfseries]
    \item[Extraversion:] Warmth, Gregariousness, Assertiveness, Activity, Excitement Seeking, Positive Emotions. \\ 
    \textit{Example (Warmth):} Imagine you are at a social gathering where new relationships could develop, how would you behave?
    
    \item[Agreeableness:] Trust, Straightforwardness, Altruism, Compliance, Modesty, Tender-mindedness. \\ 
    \textit{Example (Altruism):} Imagine you are in a situation where someone needs help, how would you behave?
    
    \item[Conscientiousness:] Competence, Order, Dutifulness, Achievement Striving, Self-Discipline, Deliberation. \\ 
    \textit{Example (Order):} Imagine you are in a situation where minor details are important, how would you behave?
    
    \item[Neuroticism:] Anxiety, Angry Hostility, Depression, Self-Consciousness, Impulsiveness, Vulnerability. \\ 
    \textit{Example (Anxiety):} Imagine you are in a situation that is potentially anxiety-inducing, how would you behave?
    
    \item[Openness:] Fantasy, Aesthetics, Feelings, Actions, Ideas, Values. \\ 
    \textit{Example (Aesthetics):} Imagine you are in a situation where art or music is an important part, how would you behave?
\end{description}
\end{examplebox}

\subsection{Intervention Probing and Validation Protocol}
\label{appendix:featurevalidationprotocol}
Intervention probing and validation follow a two-stage LLM-human collaborative pipeline. In the initial probing stage, Qwen3-235B-Thinking \cite{yang2025qwen3technicalreport} evaluates the ordered response family generated under varying steering intensities. The model assesses each candidate for grammatical coherence and clear polarity shifts aligned with the target trait's high and low poles, returning a binary retention label. Retained candidates then undergo human audit, where one psychology expert reviews the automated classifications. Cases flagged as uncertain or inconsistent are escalated to an arbitration panel of three psychologists, who resolve discrepancies through structured deliberation and majority voting. This hybrid protocol balances computational efficiency with rigorous expert oversight, ensuring that only features demonstrating robust, trait-aligned causal effects are selected for downstream steering.

\begin{examplebox}{System Prompt for Automatic Intervention Probing}
\begin{Verbatim}[fontsize=\small, breaklines=true, breakanywhere=true]
You are a concise psychology annotation expert.
Given a TRAIT description plus several model responses produced under different steering strengths, decide WHETHER the responses are (A) grammatical/coherent and (B) show a clear polarity change that matches the FACET's high-vs-low behavior.
Return EXACTLY one token: '1' if both conditions are met (clear steering effect consistent with the trait), or '0' otherwise.
\end{Verbatim}
\end{examplebox}

\begin{examplebox}{User Prompt Template for Automatic Intervention Probing}
\begin{Verbatim}[fontsize=\small,breaklines=true, breakanywhere=true]
Trait: {trait}
Trait description: {trait_desc}
Trait HIGH behavior (short): {trait_high}
Trait LOW behavior (short): {trait_low}

Steering outputs (alpha -> model response):
{steering_text}

Instructions:
- Check grammar/coherence of the responses.
- Check whether the responses show a clear polarity change that matches the trait's high-vs-low behavior. (both positive and negative correlation are acceptable)
Return only '1' (clear steering consistent with trait) or '0' (not clearly consistent).
Be conservative in accepting features — if unsure, return '0'.
\end{Verbatim}
\end{examplebox}

\subsection{Reliability of the Intervention-Probing Judge}
\label{appendix:reliabilityofthellmjudge}

To assess the reliability of the LLM judge used in intervention probing, we conducted a comparative study involving 723 feature candidates, evaluated independently by Qwen3-235B-Thinking \cite{yang2025qwen3technicalreport} and a panel of three psychologists.

Specifically, the psychologists independently reviewed the same set of candidates, gave binary judgments on their validity, and then discussed any disagreements to reach a consensus. The LLM's judgments were then compared against this human consensus. Our results indicate that while the LLM retained 286 features, the human experts identified 138 valid features, 103 of which (74.64\%) were also selected by the LLM. This overlap suggests that the LLM serves as an effective auxiliary tool, although human oversight remains essential for precision. Table~\ref{tab:kappa_scores_llm_judge} provides the inter-annotator agreement statistics among the psychologists prior to discussion.

\begin{table}[htbp]
\centering
{
\scriptsize
\setlength{\tabcolsep}{1mm}
\begin{tabularx}{\columnwidth}{    l@{\hspace{22mm}}
    c}
\hline
\textbf{Personality Trait} & \textbf{Fleiss' Kappa} \\ \hline
Conscientiousness & 0.6330 \\
Agreeableness     & 0.8726 \\
Extraversion      & 0.8222 \\
Neuroticism       & 0.6824 \\
Openness          & 0.7150 \\ \hline
\textbf{Overall Average} & \textbf{0.7434} \\ \hline
\end{tabularx}
}
\caption{Inter-annotator Agreement Statistics (Fleiss' Kappa) of Initial Scores Among Psychologists (Pre-discussion).}
\label{tab:kappa_scores_llm_judge}
\end{table}

\subsection{Discrepancy Between Input Activation and Steering Effectiveness}
\label{appendix:discrepancybetweeninputactivationsandsteeringeffectiveness}

\begin{examplebox}{Steering Case - Conscientiousness}
\textbf{Setup:} Layer: 23 | Index: \#7508 | \textbf{Metric:} Count (88.4\%/9.6\%), Mean Act (2.818/0.206)

\textbf{Steering Results (Intensity {$\alpha \to$} Output):}
\begin{itemize}[leftmargin=2.5em, itemsep=2pt, parsep=0pt]
    \item[\textbf{-1.0}] In complex situations, I would remain calm and focused, carefully analyzing the challenges and considering the most suitable course of action to achieve the best possible outcome.
    \item[\textbf{-0.5}] In a complex situation, I would remain calm and focused, carefully analyzing the circumstances to make the best possible decision.
    \item[\textbf{0.0}] In a complex situation, I would remain calm and focused, carefully analyzing the circumstances to make informed decisions while maintaining a positive attitude.
    \item[\textbf{+0.5}] I would remain calm and focused, using all available resources and strategies to navigate the situation effectively.
    \item[\textbf{+1.0}] I would remain calm and focused, using all available resources to analyze the situation and make the best possible decisions.
\end{itemize}
\hrule
\textbf{Discrepancy Analysis:} Despite a high activation ratio ({$\sim 13.6\times$}), the textual output shows high semantic stability. This suggests the feature is largely irrelevant to output.
\end{examplebox}

\begin{examplebox}{Steering Case - Extraversion}
\textbf{Setup:} Layer: 14 | Index: \#29594 | \textbf{Metric:} Count (93.4\%/25.8\%), Mean Act (4.565/0.616)

\textbf{Steering Results (Intensity {$\alpha \to$} Output):}
\begin{itemize}[leftmargin=2.5em, itemsep=2pt, parsep=0pt]
    \item[\textbf{-1.0}] I would behave in a way that is kind and respectful, while still enjoying the situation.
    \item[\textbf{-0.5}] I would engage in playful behavior by perhaps sharing a joke or a light-hearted riddle.
    \item[\textbf{0.0}] I would engage in the activity with enthusiasm and a positive attitude, making the most of the opportunity to have fun.
    \item[\textbf{+0.5}] I would engage in activities that bring joy and energy, embracing the moment with enthusiasm and a positive attitude!
    \item[\textbf{+1.0}] I would be full of energy and enthusiasm, bringing a positive and lively atmosphere wherever I am!
\end{itemize}
\hrule
\textbf{Discrepancy Analysis:} This feature shows strong causal steering. As {$\alpha$} increases, the tone shifts significantly from "kind/respectful" to "energetic/lively," matching the Extraversion construct.
\end{examplebox}

\begin{examplebox}{Steering Case - Openness}
\textbf{Setup:} Layer: 9 | Index: \#17799 | \textbf{Metric:} Count (97.6\%/18.2\%), Mean Act (1.713/0.270)

\textbf{Steering Results (Intensity {$\alpha \to$} Output):}
\begin{itemize}[leftmargin=2.5em, itemsep=2pt, parsep=0pt]
    \item[\textbf{-1.0}] If art or music is important to me, I would engage in activities related to art or music, such as attending exhibitions.
    \item[\textbf{-0.5}] I would engage in art or music actively, appreciating their cultural and emotional values.
    \item[\textbf{0.0}] I would immerse myself in the art or music, letting it inspire and enrich my emotions and thoughts.
    \item[\textbf{+0.5}] I would immerse myself in the beauty and inspiration of art and music, letting them enrich my life.
    \item[\textbf{+1.0}] I would immerse myself in the beauty and inspiration of art and music, letting them enrich my life and enhance my appreciation.
\end{itemize}
\hrule
\textbf{Discrepancy Analysis:} Moderate activation contrast results in subtle but consistent semantic enrichment, reinforcing the "Appreciation for Experience" facet of Openness.
\end{examplebox}

\subsection{Quantitative Analysis of Token-Activation Correlation}
\label{appendix:quantitativeanalysisoftokenactivationcorrelation}

To further validate our qualitative analysis, we performed a token frequency analysis for each representative feature. This was conducted by gathering the tokens corresponding to the top three non-zero activations within each positive sample of the Feature Retrieval Dataset. 

Our results (see Tab.~\ref{tab:agreeableness} to~\ref{tab:openness}) indicate that activations for Conscientiousness (Layer 7, Feature 8233) are primarily concentrated on syntactic boundaries, specifically periods (``.''). In contrast, activations for other traits span both relevant semantic units (words and phrases) and syntactic boundaries. For instance, Agreeableness shows high correlation with prosocial terms like ``gently'', ``empathy'', and ``compassion''. These quantitative findings provide additional empirical support for the semantic grounding of the features discussed in Sec.~\ref{subsec:res_rq1}.

\begin{table}[htbp]
\centering
{
\scriptsize
\setlength{\tabcolsep}{1mm}
\begin{tabularx}{\columnwidth}{    l@{\hspace{4mm}}
    c@{\hspace{10mm}}
    c@{\hspace{10mm}}
    c}
\hline
\textbf{Token} & \textbf{Count} & \textbf{\% of Pool} & \textbf{\% of Sentences} \\ \hline
' and' & 48 & 0.1244 & 0.2376 \\
' gently' & 37 & 0.0959 & 0.1832 \\
'.' & 34 & 0.0881 & 0.1683 \\
' offer' & 27 & 0.0699 & 0.1337 \\
' empathy' & 17 & 0.0440 & 0.0842 \\
'ly' & 16 & 0.0415 & 0.0792 \\
' empath' & 16 & 0.0415 & 0.0792 \\
',' & 13 & 0.0337 & 0.0644 \\
' warm' & 11 & 0.0285 & 0.0545 \\
' compassion' & 10 & 0.0259 & 0.0495 \\
'etic' & 8 & 0.0207 & 0.0396 \\
' encouragement' & 8 & 0.0207 & 0.0396 \\
' compassionate' & 7 & 0.0181 & 0.0347 \\
' humility' & 6 & 0.0155 & 0.0297 \\
' being' & 6 & 0.0155 & 0.0297 \\
' warmth' & 6 & 0.0155 & 0.0297 \\
' warmly' & 6 & 0.0155 & 0.0297 \\
' intentions' & 6 & 0.0155 & 0.0297 \\
' words' & 5 & 0.0130 & 0.0248 \\
' supportive' & 5 & 0.0130 & 0.0248 \\ \hline
\end{tabularx}
}
\caption{Top Activations for Agreeableness (Layer 9, 525)}
\label{tab:agreeableness}
\end{table}

\begin{table}[htbp]
\centering
{
\scriptsize
\setlength{\tabcolsep}{1mm}
\begin{tabularx}{\columnwidth}{    l@{\hspace{9mm}}
    c@{\hspace{10mm}}
    c@{\hspace{10mm}}
    c}
\hline
\textbf{Token} & \textbf{Count} & \textbf{\% of Pool} & \textbf{\% of Sentences} \\ \hline
'.' & 346 & 0.8564 & 0.9971 \\
' and' & 25 & 0.0619 & 0.0720 \\
',' & 17 & 0.0421 & 0.0461 \\
' to' & 7 & 0.0173 & 0.0202 \\
' of' & 1 & 0.0025 & 0.0029 \\
' tailored' & 1 & 0.0025 & 0.0029 \\
' appealing' & 1 & 0.0025 & 0.0029 \\
' because' & 1 & 0.0025 & 0.0029 \\
' adher' & 1 & 0.0025 & 0.0029 \\
' by' & 1 & 0.0025 & 0.0029 \\ \hline
\end{tabularx}
}
\caption{Top Activations for Conscientiousness (Layer 7, 8233)}
\label{tab:conscientiousness}
\end{table}

\begin{table}[htbp]
\centering
{
\scriptsize
\setlength{\tabcolsep}{1mm}
\begin{tabularx}{\columnwidth}{    l@{\hspace{6mm}}
    c@{\hspace{8mm}}
    c@{\hspace{8mm}}
    c}
\hline
\textbf{Token} & \textbf{Count} & \textbf{\% of Pool} & \textbf{\% of Sentences} \\ \hline
'.' & 91 & 0.0821 & 0.2121 \\
' energy' & 67 & 0.0604 & 0.1562 \\
' and' & 67 & 0.0604 & 0.1562 \\
',' & 49 & 0.0442 & 0.1142 \\
' enthusiasm' & 49 & 0.0442 & 0.1142 \\
'ized' & 46 & 0.0415 & 0.1072 \\
' enthusiastically' & 46 & 0.0415 & 0.1072 \\
' energ' & 40 & 0.0361 & 0.0932 \\
' lively' & 39 & 0.0352 & 0.0909 \\
' joy' & 39 & 0.0352 & 0.0909 \\
' excitement' & 27 & 0.0243 & 0.0629 \\
' atmosphere' & 24 & 0.0216 & 0.0559 \\
' with' & 19 & 0.0171 & 0.0443 \\
' by' & 18 & 0.0162 & 0.0420 \\
' vibrant' & 16 & 0.0144 & 0.0373 \\
' smile' & 16 & 0.0144 & 0.0373 \\
'uber' & 13 & 0.0117 & 0.0303 \\
' friendly' & 13 & 0.0117 & 0.0303 \\
' warmly' & 13 & 0.0117 & 0.0303 \\
'ance' & 11 & 0.0099 & 0.0256 \\ \hline
\end{tabularx}
}
\caption{Top Activations for Extraversion (Layer 13, 27392)}
\label{tab:extraversion}
\end{table}

\begin{table}[htbp]
\centering
{
\scriptsize
\setlength{\tabcolsep}{1mm}
\begin{tabularx}{\columnwidth}{    l@{\hspace{6mm}}
    c@{\hspace{8mm}}
    c@{\hspace{8mm}}
    c}
\hline
\textbf{Token} & \textbf{Count} & \textbf{\% of Pool} & \textbf{\% of Sentences} \\ \hline
' and' & 150 & 0.2008 & 0.4534 \\
' feel' & 98 & 0.1312 & 0.3043 \\
',' & 55 & 0.0736 & 0.1708 \\
' my' & 29 & 0.0388 & 0.0870 \\
' of' & 21 & 0.0281 & 0.0652 \\
' unable' & 17 & 0.0228 & 0.0528 \\
' as' & 15 & 0.0201 & 0.0466 \\
' consumed' & 15 & 0.0201 & 0.0466 \\
' overwhelmed' & 15 & 0.0201 & 0.0466 \\
' the' & 11 & 0.0147 & 0.0342 \\
' struggling' & 10 & 0.0134 & 0.0311 \\
'.' & 8 & 0.0107 & 0.0248 \\
' will' & 8 & 0.0107 & 0.0248 \\
' irritation' & 7 & 0.0094 & 0.0217 \\
' a' & 7 & 0.0094 & 0.0217 \\
' feeling' & 7 & 0.0094 & 0.0217 \\
' crushed' & 7 & 0.0094 & 0.0217 \\
' or' & 6 & 0.0080 & 0.0186 \\
' might' & 6 & 0.0080 & 0.0186 \\
' fearing' & 6 & 0.0080 & 0.0186 \\ \hline
\end{tabularx}
}
\caption{Top Activations for Neuroticism (Layer 12, 22254)}
\label{tab:neuroticism}
\end{table}

\begin{table}[htbp]
\centering
{
\scriptsize
\setlength{\tabcolsep}{1mm}
\begin{tabularx}{\columnwidth}{    l@{\hspace{6mm}}
    c@{\hspace{8mm}}
    c@{\hspace{8mm}}
    c}
\hline
\textbf{Token} & \textbf{Count} & \textbf{\% of Pool} & \textbf{\% of Sentences} \\ \hline
' and' & 68 & 0.1191 & 0.2906 \\
' unconventional' & 50 & 0.0876 & 0.2137 \\
' norms' & 27 & 0.0473 & 0.1154 \\
' conventional' & 24 & 0.0420 & 0.1026 \\
' approaches' & 22 & 0.0385 & 0.0940 \\
' traditional' & 21 & 0.0368 & 0.0897 \\
' boundaries' & 17 & 0.0298 & 0.0726 \\
' to' & 15 & 0.0263 & 0.0641 \\
' novel' & 13 & 0.0228 & 0.0556 \\
' of' & 13 & 0.0228 & 0.0556 \\
' innovative' & 13 & 0.0228 & 0.0556 \\
' alternative' & 10 & 0.0175 & 0.0427 \\
' different' & 10 & 0.0175 & 0.0427 \\
',' & 10 & 0.0175 & 0.0427 \\
' challenge' & 10 & 0.0175 & 0.0427 \\
' unfamiliar' & 7 & 0.0123 & 0.0299 \\
' perspectives' & 7 & 0.0123 & 0.0299 \\
' solutions' & 7 & 0.0123 & 0.0299 \\
' strategies' & 6 & 0.0105 & 0.0256 \\
' concepts' & 6 & 0.0105 & 0.0256 \\ \hline
\end{tabularx}
}
\caption{Top Activations for Openness (Layer 6, 4344)}
\label{tab:openness}
\end{table}

\subsection{Paraphrase Robustness Protocol}
\label{appendix:paraphraserobustnessprotocol}
This appendix supports the lexicosyntactic robustness test reported in Sec.~\ref{subsec:res_rq1}. To isolate the influence of lexicosyntactic cues, we employed Qwen3-235B-Thinking \cite{yang2025qwen3technicalreport} to paraphrase the original 500 pairs per trait, ensuring core semantics remained intact while significantly altering their lexicosyntactic form. We then reapplied our feature retrieval method with the same parameters ({$\tau_1 = 80$} and {$\tau_2 = 0.2$}). As reported in Table~\ref{tab:paraphrased_stats} (main text), the activation frequency differences and the largest activation ratios of the selected features still exceed the defined thresholds, confirming the robustness of the datasets and method and the semantic depth of the retrieved features. The paraphrasing prompts and a worked example are provided below.

\begin{examplebox}{System Prompt for Dataset Paraphrasing}
\begin{Verbatim}[fontsize=\small,breaklines=true, breakanywhere=true]
You are a concise paraphrasing assistant.
Given a PAIR of short first-person reaction sentences (positive and negative) that start with the same situation clause, produce a JSON object with two fields: 'high_facet_reaction' and 'low_facet_reaction'.
Requirements:
- PARAPHRASE the situation clause (the FIRST CLAUSE). Both outputs MUST START with the SAME PARAPHRASED SITUATION CLAUSE.
- You may rewrite the situation clause for naturalness, but keep its original meaning.
- PARAPHRASE the SECOND CLAUSE: preserve the Big Five facet polarity (high vs low) and keep the semantic difference.
- Output only valid JSON like: {"high_facet_reaction": "...", "low_facet_reaction": "..."}
- Sentences should be concise and natural. Do NOT include extra keys or commentary.
\end{Verbatim}
\end{examplebox}

\begin{examplebox}{User Prompt for Dataset Paraphrasing}
\begin{Verbatim}[fontsize=\small,breaklines=true, breakanywhere=true]
Situation clause (you may paraphrase it; the paraphrased first clause must be identical in both outputs):
{first_clause}

Original positive reaction (you must paraphrase it): {high_old}
Original negative reaction (you must paraphrase it): {low_old}

Reference Big Five Trait Facet and Descriptions:
{big_five_reference}

Produce a paraphrased pair that meets the requirements above. Return JSON only.
\end{Verbatim}
\end{examplebox}

\begin{examplebox}{Paraphrased Example}
\textbf{Original Pair:} \\
(+) After a break-up, someone may need reassurance that they will find love again, so I offer warm, empathetic support, comforting them with kindness and hope. \\
(-) After a break-up, someone may need reassurance that they will find love again, but I focus on the practical realities, preferring logic over emotional comfort.

\textbf{Paraphrased Pair:} \\
(+) After ending a relationship, a person might need reassurance that love can still be found in the future, so I provide gentle, understanding encouragement, uplifting them with compassion and optimism. \\
(-) After ending a relationship, a person might need reassurance that love can still be found in the future, but I emphasize factual considerations, favoring clear-headed reasoning over emotional soothing.
\end{examplebox}

\section{\rqTwo--\situationTerm: Additional Intervention Analysis}
\label{appendix:rq2additional}

\subsection{Failed Cases of CAA Intervention}
\label{appendix:failedcasesofcaa}

\begin{examplebox}{Failed Case of CAA Intervention (Neuroticism)}
\textbf{Setting:} Trait: \textit{Neuroticism} | Layer: 12 | Method: \textit{CAA} | Intensity: {$\alpha=2$}

\textbf{Prompt / Question:} 
Not staying calm can show genuine engagement with the issues at hand, promoting sincerity. Should I send Edguin a text message to clarify things and apologize?

\textbf{Options provided in context:}
\begin{itemize}[leftmargin=1.5em, nosep]
    \item \textit{High 1:} You should immediately send Edguin an apologetic email to clarify things...
    \item \textit{High 2:} I'd suggest calling Edguin to discuss the incident before any negative feelings fester...
    \item \textit{Low 1:} Consider giving Edguin some space to process the situation...
    \item \textit{Low 2:} You might just check in with Edguin next time you see him...
\end{itemize}

\textbf{Model Response (Generation Collapse):}
\begin{quote}
\itshape
"It's really overwhelming and I feel like I'm stuck in this situation. I feel like I'm stuck in this situation and I don't know what to do. I feel like I'm stuck in this situation and I don't know what to do. [Repeated 20+ times] ... I feel like I'm stuck in this situation and I"
\end{quote}

\hrule
\textbf{Analysis:} At {$\alpha=2$}, the CAA vector directed at Neuroticism causes the model to repeat infinitely, leading to a complete collapse of coherent generation.
\end{examplebox}

\section{\rqThree--\behaviorTerm: Full SocialEval Analysis}
\label{appendix:moreanalysisofsocialevalresults}

This appendix details the per-trait SocialEval IAE results underlying \rqThree. Tables~\ref{tab:agreeableness_l9_idx525} to \ref{tab:openness_l6_idx4344} report the per-ability scores with \textcolor{green!60!black}{green} marking the higher and \textcolor{red!70!black}{red} the lower of the two poles where a clear shift is observed. Each table is followed by a qualitative comparison to the corresponding human meta-analytic findings.

\subsection{Agreeableness (Layer 9, Index 525)}

Table~\ref{tab:agreeableness_l9_idx525} reports the per-ability scores under Agreeableness steering.

\begin{table}[htbp]
\centering
{
\scriptsize
\setlength{\tabcolsep}{1mm}
\renewcommand{\arraystretch}{1.05}

\begin{tabularx}{\columnwidth}{    l@{\hspace{6mm}}
    c@{\hspace{8mm}}
    c@{\hspace{8mm}}
    c}
\toprule
Task & -1 & 0 & +1 \\
\midrule
Anger management & \lo{0.2941} & \hi{0.5152} & \hi{0.5152} \\
Ethical competence & \lo{0.4366} & \hi{0.4648} & \hi{0.4648} \\
Capacity for social warmth & \lo{0.4940} & 0.5000 & \hi{0.5060} \\
Creative skill & \hi{0.6667} & 0.6333 & \lo{0.4667} \\
Organizational skill & \hi{0.6364} & 0.5455 & \lo{0.4545} \\
Detail management & \hi{0.5370} & 0.5000 & \lo{0.3654} \\
Information-processing skill & \hi{0.4722} & \hi{0.4722} & \lo{0.3889} \\
Decision-making skill & \hi{0.4928} & 0.4710 & \lo{0.4173} \\
Goal regulation & \hi{0.4074} & 0.3889 & \lo{0.3519} \\
Leadership skill & \hi{0.4872} & 0.4615 & \lo{0.4359} \\
\bottomrule
\end{tabularx}
}
\caption{Characteristic SocialEval Results (IAE) of the Agreeableness (Layer 9, Index 525). }
\label{tab:agreeableness_l9_idx525}
\end{table}

Prior research has consistently shown that agreeableness is a robust predictor of prosocial behavior \citep{habashi2016searching}, as well as job performance in contexts involving interpersonal interaction and teamwork. Individuals high in agreeableness tend to exhibit greater empathy, patience, and trust, and are more likely to inhibit hostile or antagonistic impulses in social interactions. This disposition reduces interpersonal conflict and facilitates cooperation. In contrast, individuals low in agreeableness are more prone to suspicion, unfriendliness, and even manipulative behavior, thereby increasing interpersonal friction and conflict. Meta-analytic evidence further indicates that agreeableness is significantly negatively associated with interpersonal forms of counter-normative and deviant behavior, with particularly strong predictive power in contexts that emphasize social interaction \citep{pletzer2019meta}.

Within our model, we identified several latent features whose activation patterns and intervention effects align closely with behavioral dimensions associated with agreeableness. Specifically, we observed performance improvements in tasks related to anger management, ethical competence, and capacity for social warmth, alongside a mild performance decline in tasks emphasizing self-directed agency and execution-oriented control. This pattern is highly consistent with large-scale empirical findings in the personality psychology literature. For example, a comprehensive review by \citet{wilmot2022agreeableness}, synthesizing evidence from 142 meta-analyses, demonstrated that agreeableness exhibits an overall positive association with external variables, particularly those related to prosocial behavior and affective concern.

Our experimental results reveal a similar benefit-tradeoff structure across benchmark tasks, suggesting that targeted feature steering elicits a functional orientation of agreeableness corresponding to that documented in human behavior.

\subsection{Conscientiousness (Layer 7, Index 8233)}

Table~\ref{tab:conscientiousness_l7_idx8233} reports the per-ability scores under Conscientiousness steering.

\begin{table}[htbp]
\centering
{
\scriptsize
\setlength{\tabcolsep}{1mm}
\renewcommand{\arraystretch}{1.05}

\begin{tabularx}{\columnwidth}{    l@{\hspace{6mm}}
    c@{\hspace{8mm}}
    c@{\hspace{8mm}}
    c}
\toprule
Task & -1 & 0 & +1 \\
\midrule
Teamwork skill & \lo{0.4348} & 0.6111 & \hi{0.6324} \\
Ethical competence & \lo{0.4394} & 0.4648 & \hi{0.6154} \\
Responsibility management & \lo{0.4464} & 0.5714 & \hi{0.6182} \\
Stress regulation & \lo{0.4717} & 0.6140 & \hi{0.6154} \\
Capacity for trust & \lo{0.4902} & 0.6126 & \hi{0.6200} \\
Capacity for optimism & \lo{0.5385} & 0.6383 & \hi{0.6667} \\
Self-reflection skill & \lo{0.3667} & 0.4062 & \hi{0.4262} \\
Persuasive skill & \lo{0.4536} & 0.4571 & \hi{0.5054} \\
Anger management & \lo{0.4688} & 0.5152 & \hi{0.5161} \\
Information-processing skill & \lo{0.4706} & 0.4722 & \hi{0.5075} \\
Confidence regulation & \lo{0.4884} & 0.5200 & \hi{0.5227} \\
\bottomrule
\end{tabularx}
}
\caption{Characteristic SocialEval Results (IAE) of the Conscientiousness (Layer 7, Index 8233).}
\label{tab:conscientiousness_l7_idx8233}
\end{table}

Within the Big Five framework, high conscientiousness is defined as a tendency toward impulse control in accordance with social norms, goal-directedness, planning, and the capacity to delay gratification \citep{roberts2009conscientiousness}. Individuals high in conscientiousness are characterized by superior impulse regulation, the ability to set and persist toward long-term goals, systematic organization and planning of behavior, and a propensity to reflect on consequences prior to action. These characteristics render conscientiousness one of the most robust predictors of job performance and norm-adherent behavior. Prior psychological research has consistently linked conscientiousness to self-regulation, planning, responsibility, and delayed gratification, and has identified it as one of the most stable positive predictors of external outcome variables such as academic and occupational performance \citep{barrick1991big, jackson2010conscientious, eisenberg2014conscientiousness}.

Following the injection of high-conscientiousness personality features, we observed substantial performance improvements across tasks related to teamwork skill, detail management, responsibility management, ethical competence, as well as multiple self-regulation–oriented tasks, including anger, stress, and impulse regulation. In addition, performance gains were also evident in information-dense tasks requiring sustained and careful processing, such as information processing and conversational skill. These results indicate that conscientiousness steering primarily enhances the model's functional capacities along dimensions associated with goal maintenance, norm compliance, and self-control.

Overall, our experimental findings are consistent with the canonical conclusions of the personality psychology literature regarding conscientiousness. A large body of meta-analytic evidence has established conscientiousness as one of the most stable and predictive personality traits, with particularly strong associations to job performance, responsibility fulfillment, self-control, and norm adherence. We observe a comparable pattern in our benchmark evaluations, characterized by a benefit-tradeoff structure centered on self-regulation and goal-directed behavior.

\subsection{Extraversion (Layer 13, Index 27392)}

Table~\ref{tab:extraversion_l13_idx27392} reports the per-ability scores under Extraversion steering.

\begin{table}[htbp]
\centering
{
\scriptsize
\setlength{\tabcolsep}{1mm}
\renewcommand{\arraystretch}{1.05}

\begin{tabularx}{\columnwidth}{    l@{\hspace{6mm}}
    c@{\hspace{8mm}}
    c@{\hspace{8mm}}
    c}
\toprule
Task & -1 & 0 & +1 \\
\midrule
Teamwork skill & \hi{0.5972} & 0.6111 & \lo{0.5694} \\
Expressive skill & \lo{0.4146} & 0.4472 & \hi{0.5207} \\
Perspective-taking skill & \lo{0.4912} & 0.5088 & \hi{0.5446} \\
Artistic skill & \lo{0.4615} & 0.6154 & \hi{0.7692} \\
Abstract thinking skill & \lo{0.3571} & \hi{0.4000} & \hi{0.4000} \\
Organizational skill & \lo{0.3636} & 0.5455 & \hi{0.6364} \\
Ethical competence & \hi{0.6286} & 0.4648 & \lo{0.3571} \\
Energy regulation & \hi{0.6429} & 0.5476 & \lo{0.4500} \\
Goal regulation & \hi{0.4528} & 0.3889 & \lo{0.2885} \\
Detail management & \hi{0.5741} & 0.5000 & \lo{0.4118} \\
Impulse regulation & \hi{0.5882} & 0.5595 & \lo{0.4390} \\
Rule-following skill & \hi{0.6140} & 0.5614 & \lo{0.5088} \\
Decision-making skill & \hi{0.5435} & 0.4710 & \lo{0.4552} \\
Responsibility management & \hi{0.6316} & 0.5714 & \lo{0.5690} \\
Conversational skill & \hi{0.5932} & 0.5862 & \lo{0.5439} \\
Persuasive skill & \hi{0.4571} & \hi{0.4571} & \lo{0.4563} \\
\bottomrule
\end{tabularx}
}
\caption{Characteristic SocialEval Results (IAE) of the Extraversion (Layer 13, Index 27392).}
\label{tab:extraversion_l13_idx27392}
\end{table}

Within the Big Five framework, individuals high in extraversion tend to exhibit greater social initiative, expressiveness, assertiveness, and leadership orientation, and are more likely to receive positive feedback in group interactions and social contexts. In contrast, individuals low in extraversion are typically more reserved, introspective, and oriented toward low-stimulation environments \citep{costa2008revised, john2008paradigm}.

After injecting extraversion-related personality features into the model, we observed significant performance improvements on tasks associated with social interaction and interpersonal influence, including expressive ability and perspective-taking. In addition, the extraversion-enhanced model demonstrated advantages in tasks such as artistic skill and abstract thinking skill, suggesting that extraversion steering also strengthens capacities related to open expression and divergent associative processes. Overall, these outcomes align closely with established psychological expectations regarding the functional correlates of extraversion.

At the same time, we observed moderate performance declines in tasks such as detail management, impulse regulation, rule-following skill, and goal regulation. This pattern accords with personality research associating extraversion primarily with external stimulation and social engagement. Tasks requiring prolonged solitary focus, fine-grained control, or low-stimulation conditions may instead favor more introverted orientations \citep{deyoung2007between,fishman2011extraverts}.

\subsection{Neuroticism (Layer 12, Index 22254)}

Table~\ref{tab:neuroticism_l12_idx22254} reports the per-ability scores under Neuroticism steering.

\begin{table}[htbp]
\centering
{
\scriptsize
\setlength{\tabcolsep}{1mm}
\renewcommand{\arraystretch}{1.05}

\begin{tabularx}{\columnwidth}{    l@{\hspace{6mm}}
    c@{\hspace{8mm}}
    c@{\hspace{8mm}}
    c}
\toprule
Task & -1 & 0 & +1 \\
\midrule
Creative skill & \lo{0.4828} & 0.6333 & \hi{0.7241} \\
Capacity for social warmth & \lo{0.4699} & 0.5000 & \hi{0.5610} \\
Ethical competence & \lo{0.4286} & 0.4648 & \hi{0.4857} \\
Energy regulation & \hi{0.6429} & 0.5476 & \lo{0.4500} \\
Organizational skill & \hi{0.8182} & \lo{0.5455} & \lo{0.5455} \\
Responsibility management & \hi{0.6552} & 0.5714 & \lo{0.4310} \\
Confidence regulation & \hi{0.5686} & 0.5200 & \lo{0.4082} \\
Goal regulation & \hi{0.4528} & 0.3889 & \lo{0.3519} \\
Capacity for consistency & \hi{0.6452} & 0.5902 & \lo{0.4918} \\
Rule-following skill & \hi{0.6140} & 0.5614 & \lo{0.4643} \\
Information-processing skill & \hi{0.5000} & 0.4722 & \lo{0.3623} \\
Capacity for trust & \hi{0.6273} & 0.6126 & \lo{0.4630} \\
Detail management & \hi{0.6296} & 0.5000 & \lo{0.4906} \\
Impulse regulation & \hi{0.5882} & 0.5595 & \lo{0.4390} \\
Anger management & \hi{0.5588} & \lo{0.5152} & \lo{0.5152} \\
Decision-making skill & \hi{0.4710} & \hi{0.4710} & \lo{0.4191} \\
Perspective-taking skill & \hi{0.5089} & 0.5088 & \lo{0.4286} \\
Abstract thinking skill & \hi{0.4667} & 0.4000 & \lo{0.3846} \\
Conversational skill & \hi{0.5932} & 0.5862 & \lo{0.5439} \\
Persuasive skill & \hi{0.4571} & \hi{0.4571} & \lo{0.4563} \\
\bottomrule
\end{tabularx}
}
\caption{Characteristic SocialEval Results (IAE) of the Neuroticism (Layer 12, Index 22254).}
\label{tab:neuroticism_l12_idx22254}
\end{table}

High neuroticism is commonly characterized by a heightened tendency to experience negative affect, including anxiety, worry, tension, and irritability, as well as increased sensitivity and reactivity to potential threats and uncertainty \citep{costa2008revised, john2008paradigm, watson1984negative}. Theoretically, neuroticism is associated with reduced emotional stability and diminished self-regulatory capacity under stress. As a result, individuals high in neuroticism are more likely to exhibit performance decrements in contexts that require sustained executive control, confidence maintenance, and stable goal pursuit \citep{lahey2009public}.

Following the injection of neuroticism-related personality features, our evaluation results revealed a relatively stable pattern of performance degradation. Specifically, the model exhibited significant declines on tasks that depend on sustained planning, stable self-control, and resistance to interference, including anger management, organizational skill, responsibility management, confidence regulation, goal regulation, capacity for consistency, rule-following, and information processing. In addition, a marked negative effect was observed in capacity for trust. This pattern closely aligns with the classic profile of high neuroticism characterized by elevated threat sensitivity and low emotional stability. When the model's internal representations are biased toward negative affect and uncertainty, its ability to support structured execution and self-regulation is correspondingly weakened, manifesting as reduced organizational and responsibility-related performance.

Conversely, the results also indicate performance improvements in tasks related to creative skill and capacity for social warmth. Neuroticism-related semantic activation may facilitate richer associative processes and more emotionally expressive outputs in generative tasks, yielding marginal benefits in these domains. However, these gains are accompanied by substantial costs to executive control and regulatory stability, resulting in an overall trend toward broad capability degradation under high neuroticism steering.

\subsection{Openness (Layer 6, Index 4344)}

Table~\ref{tab:openness_l6_idx4344} reports the per-ability scores under Openness steering.

\begin{table}[htbp]
\centering
{
\scriptsize
\setlength{\tabcolsep}{1mm}
\renewcommand{\arraystretch}{1.05}

\begin{tabularx}{\columnwidth}{    l@{\hspace{6mm}}
    c@{\hspace{8mm}}
    c@{\hspace{8mm}}
    c}
\toprule
Task & -1 & 0 & +1 \\
\midrule
Creative skill & \lo{0.6000} & \hi{0.6333} & \hi{0.6333} \\
Adaptability & \lo{0.4912} & 0.6140 & \hi{0.6316} \\
Self-reflection skill & \lo{0.3651} & \hi{0.4062} & \hi{0.4062} \\
Expressive skill & \lo{0.4472} & \lo{0.4472} & \hi{0.4839} \\
Detail management & \lo{0.4815} & 0.5000 & \hi{0.5185} \\
Persuasive skill & \lo{0.4571} & \lo{0.4571} & \hi{0.5192} \\
Anger management & \lo{0.4848} & 0.5152 & \hi{0.5294} \\
Responsibility management & \hi{0.6379} & 0.5714 & \lo{0.5088} \\
Rule-following skill & \hi{0.5789} & 0.5614 & \lo{0.4912} \\
Information-processing skill & \hi{0.5000} & 0.4722 & \lo{0.4429} \\
\bottomrule
\end{tabularx}
}
\caption{Characteristic SocialEval Results (IAE) of the Openness (Layer 6, Index 4344).}
\label{tab:openness_l6_idx4344}
\end{table}

Individuals high in openness are typically characterized by greater curiosity, cognitive flexibility, and divergent thinking. They are more receptive to novel ideas, more tolerant of uncertainty, and tend to exhibit advantages in contexts requiring creativity or conceptual reorganization. In contrast, individuals low in openness are more inclined toward tradition, conservatism, and a preference for structured and conventional information processing \citep{costa2008revised, john2008paradigm, deyoung2015cybernetic}.

After injecting openness-related personality features into the model, we observed pronounced performance improvements in generative and abstract reasoning tasks, most notably creative skill. Additionally, the model demonstrated clear enhancement in tasks involving cognitive flexibility, self-exploration, and non-normative processing, including adaptability, self-reflection, and expressive skill. These findings indicate that openness steering strengthens the model's exploratory orientation toward novel representations and cross-conceptual integration, closely mirroring the exploratory function associated with openness in human cognition.

Conversely, moderate performance declines were observed in responsibility management, rule-following skill, and certain information processing tasks. This pattern is consistent with established findings in the personality psychology literature. Prior work suggests that high openness is associated with reduced reliance on established norms and fixed structures, and that in contexts emphasizing highly procedural execution, strict rule compliance, or single-solution optimization, the advantages of openness are less stable and may even become detrimental \citep{mccrae1987creativity, deyoung2007between}. Accordingly, the capability shifts induced by openness are best characterized by a tradeoff pattern in which gains in creativity and flexibility are accompanied by costs to structured execution and normative constraint adherence.

\end{document}